%% file: main.tex
\documentclass[10pt,twocolumn,letterpaper]{article}
\usepackage{iftex}
\ifPDFTeX\pdfminorversion=7\fi
\ifXeTeX\fi
\usepackage[pagenumbers]{cvpr}
\usepackage{xspace}
\usepackage{newunicodechar}
\newunicodechar{’}{\textquoteright}
\newunicodechar{…}{\ldots}
\newcommand{\baby}{ICML\xspace}
\newcommand{\paratitle}[1]{\vspace{1.5ex}\noindent \textbf{#1}}
\newcommand{\eat}[1]{}
\input{preamble}
\usepackage[breaklinks,hidelinks,pdfusetitle]{hyperref}

\newcommand{\arxivcodeavailability}{The code is publicly available at \url{https://github.com/Ly041021/ICML}.}
\title{Trust to Teach: Intra-IMU Spectral Encoding and Cross-IMU Mutual Learning for Inertial Odometry}
\input{arxiv_authors}
\date{}

\begin{document}
\maketitle
\input{sec/0_abstract}    
\input{sec/teaser}
\input{sec/1_intro}

\input{sec/2_related_work}
\input{sec/3_method}
\input{sec/4_Experiments}
\input{sec/5_Conclusion}
\input{sec/6_llm_usage}

{
    \small
    \bibliographystyle{ieeenat_fullname}
    \bibliography{main}
}

\end{document}

%% file: preamble.tex
\usepackage[T1]{fontenc}
\usepackage{multirow}
\usepackage{colortbl}
\definecolor{IOHeader}{RGB}{237,241,244}
\definecolor{IOPrimary}{RGB}{229,234,241}
\definecolor{IOBranch}{RGB}{247,248,250}
\definecolor{IOFocus}{RGB}{13,47,87}
\definecolor{IOGain}{RGB}{13,66,134}
\definecolor{IOLoss}{RGB}{160,45,42}
\definecolor{IORule}{RGB}{180,187,196}
\newif\ifIOFullStats
\newcommand{\IOstd}[1]{\ifIOFullStats\,{\scriptstyle\pm\,#1}\fi}
\newcommand{\IOgroupRule}{\arrayrulecolor{IORule}\midrule[.25pt]\arrayrulecolor{black}}

\newcommand{\IOvalue}[1]{\ifIOFullStats#1\else\makebox[30pt][r]{#1}\fi}
\newcommand{\IOdeltaBlank}{\ifIOFullStats\else\hspace{3pt}\makebox[30pt]{}\fi}
\newcommand{\IOdelta}[2]{%
  \ifIOFullStats
    \hspace{1.5pt}\raisebox{1pt}[0pt][0pt]{\makebox[26pt][r]{\fontsize{6.5}{7}\selectfont\bfseries\textcolor{#1}{(#2\%)}}}%
  \else
    \hspace{3pt}\raisebox{0.7pt}[0pt][0pt]{\makebox[30pt][r]{\fontsize{7}{8}\selectfont\bfseries\textcolor{#1}{(#2\%)}}}%
  \fi
}

\input{tables/ours_reference_bindings}

\usepackage{stfloats}


\usepackage{tikz}
\usetikzlibrary{arrows.meta,positioning,fit}

\usepackage{microtype}



%% file: tables/ours_reference_bindings.tex
\newcommand{\IOBindResult}[4]{%
  \ifIOFullStats\else\expandafter\gdef\csname IOResult@#1@#2@#3\endcsname{#4}\fi%
  #4%
}
\DeclareRobustCommand{\IOResult}[3]{%
  \begingroup
  \IOFullStatstrue
  \renewcommand{\IOvalue}[1]{##1}%
  \renewcommand{\IOdelta}[2]{}%
  \let\IOdeltaBlank\relax
  \ifcsname IOResult@#1@#2@#3\endcsname
    \csname IOResult@#1@#2@#3\endcsname
  \else
    \PackageError{IOresults}{Missing Table 1 result #1/#2/#3}{Typeset Table 1 before its dependent tables.}%
  \fi
  \endgroup
}

\newcommand{\IORegisterResult}[4]{\expandafter\gdef\csname IOResult@#1@#2@#3\endcsname{#4}}

%% file: arxiv_authors.tex
\author{%
\begingroup\small
\renewcommand{\arraystretch}{0.89}
\begin{tabular}[t]{@{}c@{}}
{\normalsize
Yan Li\textsuperscript{\ensuremath{\dagger}1},\enspace
Yang Liu\textsuperscript{\ensuremath{\dagger}1},\enspace
Yuanchao Shu\textsuperscript{1},\enspace
Jiming Chen\textsuperscript{1},\enspace
Liang Li\textsuperscript{*1}}\\[0.05em]
{\small \textsuperscript{1}Zhejiang University, Hangzhou, China}\\
{\small \textsuperscript{\ensuremath{\dagger}}These authors contributed equally to this work.}\\
{\small \textsuperscript{*}Corresponding author. E-mail: liang.li@zju.edu.cn (Liang Li).}\relax
\end{tabular}
\endgroup%
}
\hypersetup{pdfauthor={Yan Li, Yang Liu, Yuanchao Shu, Jiming Chen, Liang Li}}

%% file: sec/0_abstract.tex
\begin{abstract}
Neural inertial odometry typically estimates motion from a single inertial measurement unit (IMU). On wearable systems and embodied platforms, IMUs at different locations capture complementary cues to body motion, allowing a velocity estimate from one IMU to serve as a training signal for the other. The reliability of this signal, however, varies with local motion, and either sensor stream may be interrupted. The central challenge is to derive reliable cross-IMU supervision from observations of varying quality while retaining the ability to estimate body motion from either IMU alone. To this end, we introduce \baby, a dual-IMU odometry framework that combines \textbf{I}ntra-IMU spectral encoding with \textbf{C}ross-IMU \textbf{M}utual \textbf{L}earning. Within each branch, a Local Dynamics Expert (LDE) captures multiscale motion changes from wavelet subbands, while a Global Spectral Expert (GSE) models cross-frequency dependencies in the full-window Fourier spectrum. A learned gate fuses these representations, enabling each branch to independently estimate the velocity of a common body reference point and its componentwise uncertainty. For each synchronized window, we jointly model the branches’ predictive variances and error correlation to derive minimum-variance fusion weights for each velocity component. These weights fuse the estimates at inference and determine how strongly each branch teaches its peer during training, allowing either branch to take the lead as motion and relative reliability vary across axes. Experiments on RoNIN, CoWear, GrandTour, and a self-collected UAV dataset span wearable, aerial, and quadruped motion. Against the strongest reported single-IMU baselines, \baby reduces ATE by 12.4--30.5\% across the four datasets.
\ifdefined\arxivcodeavailability\arxivcodeavailability\fi
\end{abstract}

%% file: sec/teaser.tex
\begin{figure}[t]
    \centering
    \includegraphics[width=\columnwidth]{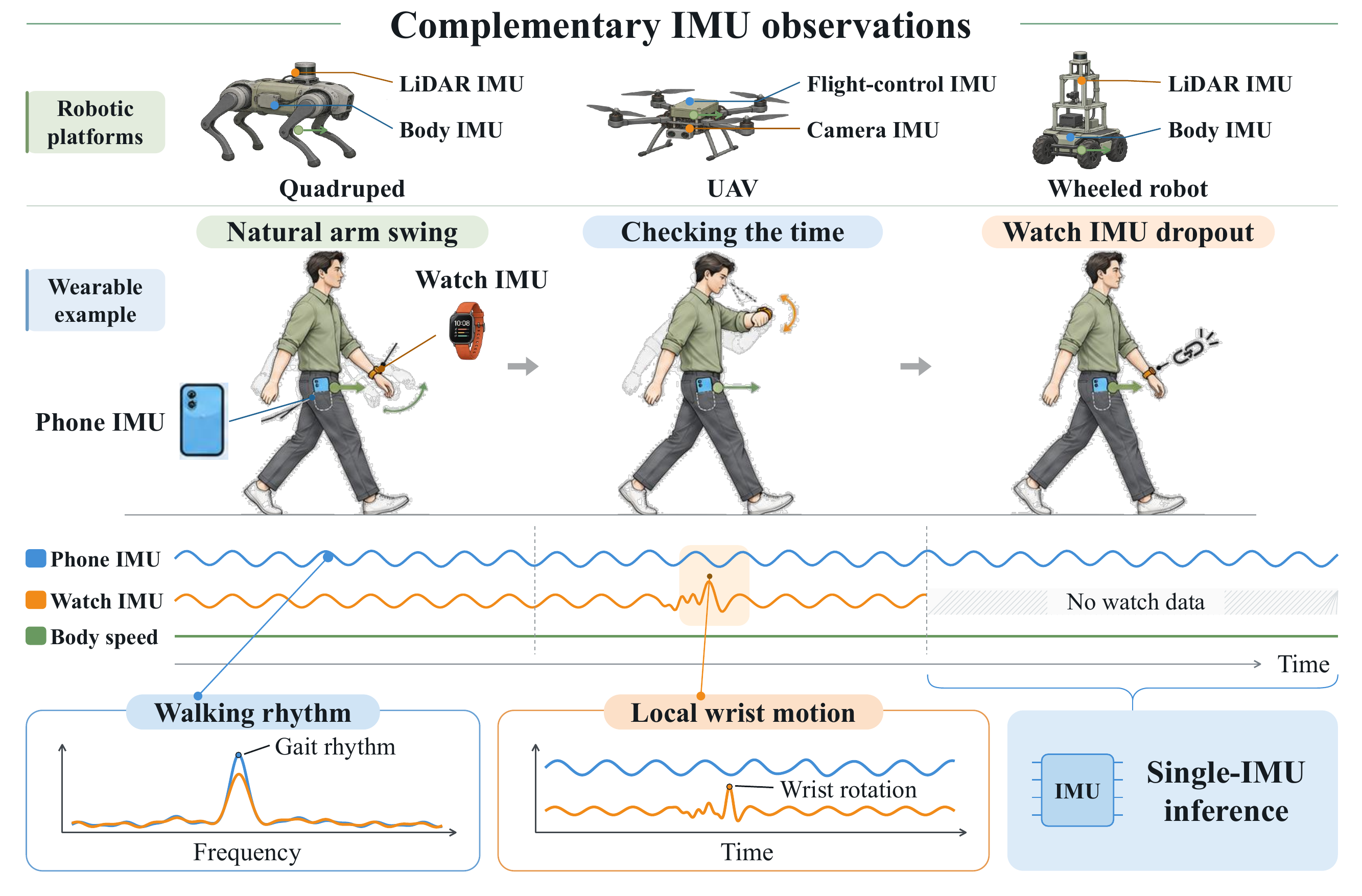}
    \caption{\textbf{Complementary IMU observations.} The top row shows example IMU placements on robotic platforms. In the wearable example, checking the time changes the watch signal while walking speed stays constant. Training uses paired streams, and inference continues from the phone alone during watch dropout. Signals and sensor locations are schematic.}
    \label{fig:teaser}
\end{figure}

%% file: sec/1_intro.tex
\section{Introduction}
\label{sec:intro}
The inertial measurement unit (IMU) plays a key role in sensing self-motion, much as the vestibular system senses head motion in humans. Inertial odometry uses high-rate, three-axis accelerometer and gyroscope measurements to estimate body velocity and relative displacement without external references. These estimates support pedestrian tracking~\cite{feliz2009pedestrian} and autonomous robot navigation~\cite{gul2019comprehensive} in settings with limited satellite coverage, such as indoor search and rescue and underground inspection~\cite{protopapadakis2016autonomous}.

Most inertial odometry methods take measurements from a single IMU as input and broadly fall into two categories: conventional inertial dead reckoning and data-driven neural inertial tracking. Conventional methods estimate motion using kinematic models or application-specific priors. Strapdown inertial navigation~\cite{titterton2004strapdown} propagates attitude from gyroscope measurements and integrates gravity-compensated acceleration to estimate velocity and displacement. Even small sensor biases or attitude errors, however, can accumulate through integration and cause substantial position drift. Pedestrian dead reckoning~\cite{han2023waveform,tian2015enhanced} instead estimates displacement from step length and heading, but irregular gait can undermine step detection and step-length estimation. 

These limitations motivate learning the relationship between inertial measurements and body motion directly from data. Neural inertial tracking methods~\cite{yan2020ronin,chen2018ionet,li2025m2eit,rao2022ctin,li2025minf,zeinali2024imunet,wang2022a2dio,li2025lite} learn to predict velocity or relative displacement from short IMU sequences, capturing motion patterns that are difficult to model by hand. Despite these advances, three challenges remain in learning from inertial observations: 1) Most methods estimate body motion from a single IMU, leaving complementary observations from other locations unused. 2) Across IMUs, the relative reliability of motion estimates varies over time, and individual data streams may be interrupted. This makes it difficult to determine when one estimate can reliably guide another. 3) Local time--frequency patterns and the spectral structure of the full observation window offer complementary cues to body motion, yet their joint use remains underexplored. 

To illustrate these points, we give an example in Figure~\ref{fig:teaser}. First, pedestrians and mobile platforms can carry two IMUs at different locations. For a pedestrian, one IMU is in the phone in their pocket and the other is in the watch on their wrist. On a drone or a legged robot, the two IMUs can be mounted at different points. Second, when a pedestrian raises their wrist to check the time without changing speed, the watch records the brief wrist rotation, while the phone may better capture the walking rhythm. Therefore, it becomes important to investigate the impact of cooperation between two IMUs on inertial odometry.

To this end, we introduce a dual-IMU odometry framework (called \baby) that combines \textbf{I}ntra-IMU spectral encoding with \textbf{C}ross-IMU \textbf{M}utual \textbf{L}earning. The central idea is that either branch can take the lead as teacher, depending on the observation window and the velocity component being estimated. First, we design Dual-Domain Spectral Experts to capture both local motion dynamics and the spectral structure of the full window. The Local Dynamics Expert uses wavelet sub-bands to capture motion changes at multiple time scales, while the Global Spectral Expert models cross-frequency dependencies in the Fourier spectrum of the entire window. A spectral fusion gate integrates these complementary features within each IMU branch, allowing each branch to independently predict the velocity of a common body reference point and the uncertainty of each velocity component. Second, we introduce a Cross-IMU Fusion and Mutual Learning module that uses the same weights to fuse velocity estimates at inference and to guide mutual teaching during training. For correlation-aware fusion, we jointly model each branch's predictive variance and the correlation between the two branches' errors. Within each window, we combine the two velocity estimates component by component using weights that minimize the predicted variance of each fused component. For directional mutual learning, we use each branch's relative contribution to fusion to determine how strongly its prediction supervises its peer. Each branch can thus learn from motion cues at the other sensor location during paired training while retaining the ability to estimate body velocity from its own measurements at inference. We evaluate \baby on RoNIN, CoWear, GrandTour, and a self-collected UAV dataset, covering human, aerial, and quadruped motion, with controlled studies examining observation fusion and mutual learning.

Our main contributions are as follows:
\begin{itemize}

\item We introduce a dual-IMU odometry framework (named \baby ) that uses each branch's contribution to velocity fusion to determine its influence as a teacher. By modeling the correlation between the branches' prediction errors, we derive fusion weights that let the two IMUs exchange teaching roles across observation windows and velocity components while retaining independent inference.

\item We design \emph{Dual-Domain Spectral Experts} to capture complementary motion cues within each IMU stream. A Local Dynamics Expert mixes temporal patches within wavelet subbands, while a Global Spectral Expert mixes frequency patches across the Fourier spectrum. A learned gate combines these representations to estimate a common body velocity and its componentwise uncertainty.

\item We distinguish the benefit of mutual learning from that of a second sensor at inference by evaluating jointly and independently trained branches under identical single-IMU inputs. Experiments across four datasets assess odometry accuracy and single-branch inference.

\end{itemize}

%% file: sec/2_related_work.tex
\section{Related Work}
\label{sec:related_work}

\subsection{Traditional Inertial Navigation}
Early approaches to inertial tracking drew on Newtonian mechanics to reconstruct body motion from onboard measurements. A classical example is the strapdown inertial navigation system~\cite{kong2000inertial,garg1978strapdown}, which uses gyroscope readings to propagate attitude, transforms measured specific force into the navigation frame, and integrates gravity-compensated acceleration to estimate velocity and position~\cite{savage1998strapdown,titterton2004strapdown}. Although SINS requires no external infrastructure, even a small tilt error can cause gravity to appear as horizontal acceleration. Together with sensor bias, this error leads to accumulating position drift.

To limit drift, classical pedestrian tracking methods exploit the structure of human gait. Pedestrian dead reckoning (PDR) updates position from estimated step length and heading~\cite{han2023waveform,tian2015enhanced}, while foot-mounted systems apply zero-velocity updates (ZUPTs) during stationary intervals of the stance phase~\cite{foxlin2005pedestrian,park2010zupt}. Irregular gait can compromise step-length estimation, and changes in sensor placement can invalidate the zero-velocity assumption, making these methods difficult to apply reliably in everyday settings.

\subsection{Learning-Based Inertial Odometry}
In recent years, learning-based inertial tracking~\cite{li2025implicit,yan2020ronin,li2025minf} has used motion cues in short IMU sequences to mitigate integration drift. Early methods such as IONet~\cite{chen2018ionet} and RoNIN~\cite{yan2020ronin} directly regressed displacement or velocity from inertial windows. TLIO~\cite{liu2020tlio} extended this approach by incorporating predicted 3D displacements and their uncertainties into an extended Kalman filter for state estimation.

Subsequent work has advanced inertial representation learning along two directions. Gravity-aligned canonicalization and rotational equivariance address changes in sensor orientation~\cite{cao2022rio,jayanth2025eqnio,kim2026ginio}, while TartanIMU~\cite{zhao2025tartanimu} and AirIO~\cite{qiu2025airio} address platform differences through cross-platform pretraining and attitude-aware modeling of aerial motion, respectively. A complementary line of work enriches IMU representations by capturing temporal context through attention~\cite{rao2022ctin} and disentangling and aggregating multiscale motion patterns across domains through a mixture of experts~\cite{li2025m2eit}. Despite these advances, tracking accuracy with a single IMU remains limited. 

Recent research has begun to explore collaborative tracking with multiple IMUs. CoWear~\cite{she2026cowear}, for example, trains device-specific estimators to predict the displacement of a common reference device and fuses their predictions using learned uncertainty estimates. However, how IMU branches at different locations can learn cooperatively during training remains underexplored.

%% file: sec/3_method.tex
\section{Method}
\label{sec:method}

\begin{figure*}[t]
    \centering
    \includegraphics[width=\textwidth]{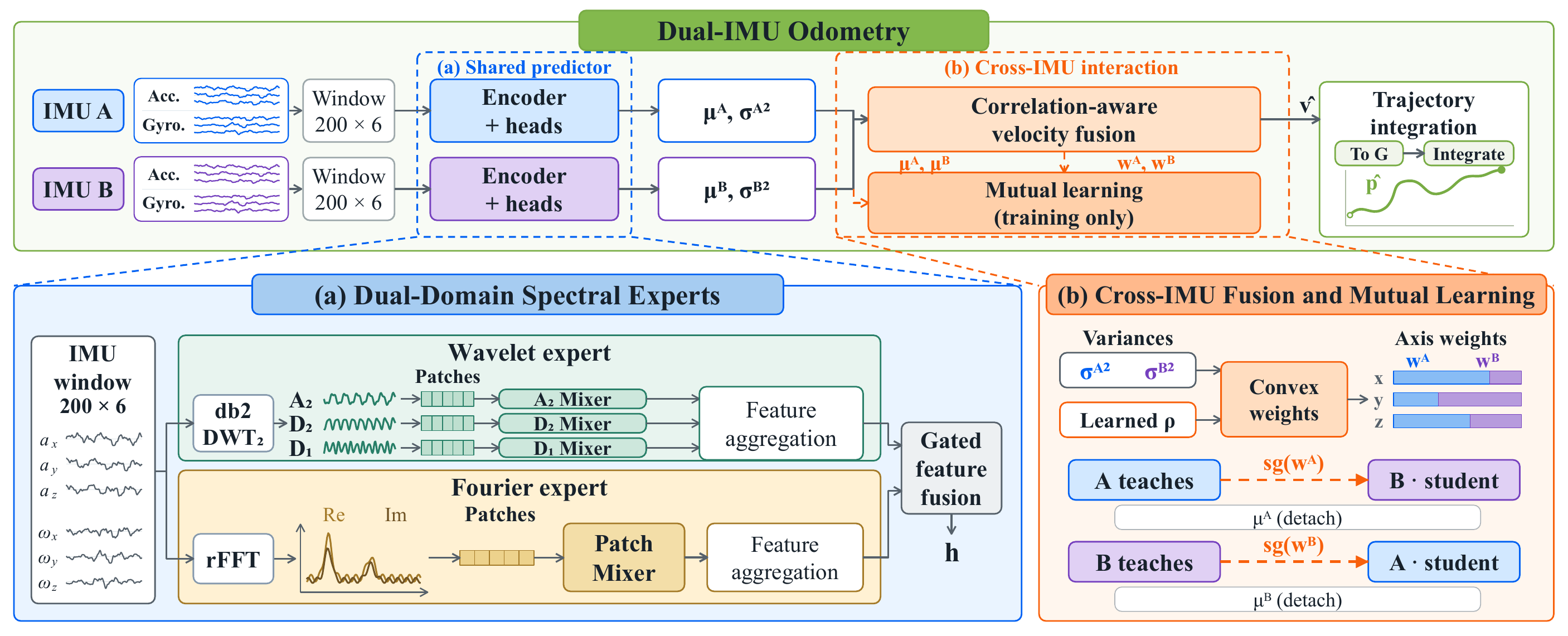}
    \caption{\textbf{Overview of \baby.} Two IMUs share a spectral encoder and heads for common-target velocity and axis-wise variance. (a) Gated fusion combines wavelet and Fourier features. (b) Predicted variances and learned error correlations yield convex weights for inference fusion and asymmetric training-time teaching. Teaching detaches peer predictions and weights; fused velocities are integrated into trajectories.}
    \label{fig:framework}
\end{figure*}

Figure~\ref{fig:framework} provides an overview of \baby. First, this section formulate dual-IMU odometry as estimating the velocity of a common body reference point from synchronized inertial observations at two locations (Sec.~\ref{sec:problem}). Next, we introduce the dual-domain spectral experts, which form a shared encoder that allows each branch to predict this velocity and its componentwise uncertainty from its own observations (Sec.~\ref{sec:spectral}). We then describe cross-IMU fusion and mutual learning, where the covariance of the two branches' prediction errors determines how much each estimate contributes to fusion and how strongly it guides its peer during training (Sec.~\ref{sec:reliability}). Finally, we present the overall training objective (Sec.~\ref{sec:training}).

\subsection{Task Formulation}
\label{sec:problem}

Consider two synchronized IMUs indexed by $k\in\{A,B\}$, each providing a window of six-axis inertial measurements $\mathbf{X}_n^k\in\mathbb{R}^{N\times 6}$ (specific force and angular velocity) at timestamp $t_n$. Although the two sensors experience distinct local kinematics due to their mounting locations or relative articulation, both remain anchored to the same moving agent. We therefore formulate both branches to estimate the velocity of a shared reference point, which absorbs relative inter-sensor motion without rigid-body lever-arm calibration and aligns both predictions in a common space for cross-sensor supervision.

Let $G$ denote a fixed inertial reference frame and $F_n$ denote the prediction frame (either $G$ or a common body frame). Over the preceding interval $\tau$, the reference velocity target is
\begin{equation}
\mathbf{v}_n^{F,*}
=(\mathbf{R}_n^{G\leftarrow F,*})^\top
\frac{\mathbf{p}^*(t_n)-\mathbf{p}^*(t_n-\tau)}{\tau},
\label{eq:target}
\end{equation}
where $\mathbf{p}^*$ denotes the reference position trajectory. Ground-truth orientations $\mathbf{R}_n^{G\leftarrow F,*}$ are used exclusively to construct training targets and losses, whereas deployment relies solely on estimated attitudes.

Both branches employ a Siamese predictor with shared parameters $\theta$:
\begin{equation}
(\boldsymbol{\mu}_n^k,\mathbf{s}_n^k)=f_\theta(\mathbf{X}_n^k),
\qquad \boldsymbol{\sigma}_n^{k,2}=\exp(\mathbf{s}_n^k),
\label{eq:prediction}
\end{equation}
where $\boldsymbol{\mu}_n^k\in\mathbb{R}^3$ is the predicted reference velocity and $\mathbf{s}_n^k\in\mathbb{R}^3$ parameterizes the bounded axis-wise log-variance. Sharing $f_\theta$ without sensor-identity embeddings applies a unified motion predictor to both streams, preserving input-dependent predictive differences while enabling each branch to estimate motion independently.

\subsection{Dual-Domain Spectral Experts}
\label{sec:spectral}
To predict accurate velocities and well-calibrated uncertainties across varying local dynamics, the shared encoder $f_\theta$ must distinguish sustained body-motion rhythms from transient local disturbances. In inertial measurements, local shifts in subband energy capture transient events such as wrist gestures or step impacts, whereas dominant frequencies across the full window characterize sustained rhythms such as walking cadence or rotor vibrations. We capture both regimes within each IMU branch by pairing a wavelet-based Local Dynamics Expert (LDE) with a Fourier-based Global Spectral Expert (GSE).

\paratitle{Local dynamics expert.}
LDE applies a two-level \texttt{db2} discrete wavelet transform (DWT) to decompose each inertial channel into one approximation subband and two detail subbands $\{\mathbf{A}_2,\mathbf{D}_2,\mathbf{D}_1\}$. Following patch mixing~\cite{murad2025wpmixer}, each subband is embedded independently as $\mathbf{Z}_0\in\mathbb{R}^{6\times P\times C}$ with $P$ temporal patches and $C$ latent features, and processed by two mixing stages:
\begin{equation}
\begin{aligned}
\mathbf{U}_\ell&=\mathbf{Z}_{\ell-1}
+\mathcal{T}_\ell(\operatorname{LN}(\mathbf{Z}_{\ell-1})),\\
\mathbf{Z}_\ell&=\mathcal{A}_\ell\!\left(
\mathbf{U}_\ell+\mathcal{F}_\ell(\operatorname{LN}(\mathbf{U}_\ell))\right),
\end{aligned}
\label{eq:spectral_mixing}
\end{equation}
for $\ell\in\{1,2\}$. Here, $\mathcal{T}_\ell$ and $\mathcal{F}_\ell$ mix temporal patches and latent features, while $\mathcal{A}_\ell$ applies residual mixing across the six accelerometer and gyroscope axes to model inter-axis kinematic coupling, with each operator instantiated as a two-layer GELU MLP. An outer residual yields $\bar{\mathbf{Z}}=\operatorname{LN}(\mathbf{Z}_1+\mathbf{Z}_2)$. For each subband, we average global mean pooling with the last complete temporal patch to anchor window-level context to the terminal prediction interval $\tau$, then concatenate and project the three subband readouts into $\mathbf{h}_D^k\in\mathbb{R}^{128}$.

\paratitle{Global spectral expert.}
In parallel, an rFFT along the temporal axis computes the one-sided complex spectrum $\mathbf{S}^k=\operatorname{rFFT}(\mathbf{X}_n^k)$, packing real and imaginary components into frequency patches to preserve phase information. A dedicated encoder applies the same two-stage mixing architecture (Eq.~\eqref{eq:spectral_mixing}), with $\mathcal{T}_\ell$ operating across frequency patches. Global average pooling over frequency patches followed by linear projection produces $\mathbf{h}_F^k\in\mathbb{R}^{128}$.

Finally, we design a spectral fusion gate (SFG) that adaptively fuses wavelet and Fourier features by jointly weighting their contributions as follows:
\begin{equation}
\begin{aligned}
\boldsymbol{\alpha}^k
&=\operatorname{softmax}(g_\phi([\mathbf{h}_D^k;\mathbf{h}_F^k])),\\
\mathbf{h}^k
&=\operatorname{MLP}([\alpha_D^k\mathbf{h}_D^k;\alpha_F^k\mathbf{h}_F^k]),
\end{aligned}
\label{eq:spectral_fusion}
\end{equation}
where $\boldsymbol{\alpha}^k=[\alpha_D^k,\alpha_F^k]^\top\in\mathbb{R}^2$. Following a shared linear projection, $\mathbf{h}^k$ feeds the velocity head directly and passes through a stop-gradient path to the variance head, isolating uncertainty calibration from motion representation learning.

\subsection{Cross-IMU Fusion and Mutual Learning}
\label{sec:mutual}
\label{sec:reliability}

Although the two IMU branches estimate the velocity of the same body reference point, their relative reliability varies with motion dynamics and across velocity components. We leverage this variation to guide mutual learning: for each window and velocity component, a branch's contribution to the fused estimate determines how strongly its prediction supervises its peer. Both branches can thus learn from paired observations while retaining the ability to predict velocity independently.

\paratitle{Correlation-aware fusion.}
Because the branches observe coupled motion and share a predictor $f_\theta$, their estimation errors can be correlated. To account for this dependence, we model the joint prediction error separately for each window $n$ and velocity component $d\in\{1,2,3\}$. Let $a=\sigma_{n,d}^{A,2}$ and $b=\sigma_{n,d}^{B,2}$ denote the marginal predictive variances. The corresponding error covariance is
\begin{equation}
\boldsymbol{\Sigma}_{n,d}
=
\begin{bmatrix}
a & q \\
q & b
\end{bmatrix},
\qquad
q=\rho_{n,d}\sqrt{ab}.
\label{eq:covariance}
\end{equation}
A correlation head predicts $\boldsymbol{\rho}_n=0.95\tanh(g_\psi(\mathbf{c}_n))$, where $\mathbf{c}_n$ summarizes gradient-isolated window statistics, inter-branch prediction differences, and log-variances using permutation-invariant sums and absolute differences. The predicted correlations are therefore invariant when the two IMUs are exchanged, while the bound $|\rho_{n,d}|<0.95$ ensures that $\boldsymbol{\Sigma}_{n,d}$ remains positive definite for all $a,b>0$.

Let $\mathbf{w}_{n,d}=[w_{n,d}^A,w_{n,d}^B]^\top$ denote nonnegative weights summing to one. The fused velocity and its predictive variance are
\begin{equation}
\begin{aligned}
\hat{v}_{n,d}
&=w_{n,d}^A\mu_{n,d}^A+w_{n,d}^B\mu_{n,d}^B,\\
\sigma_{f,n,d}^2
&=\mathbf{w}_{n,d}^\top\boldsymbol{\Sigma}_{n,d}\mathbf{w}_{n,d}.
\end{aligned}
\label{eq:fused_prediction}
\end{equation}
Minimizing this variance over admissible convex weights yields
\begin{equation}
w_{n,d}^A
=\operatorname{clip}\!\left(
\frac{b-q}{a+b-2q},\,0,\,1
\right),
\qquad
w_{n,d}^B=1-w_{n,d}^A.
\label{eq:fusion_weights}
\end{equation}
When $\rho_{n,d}=0$, this rule reduces to inverse-variance weighting; when only one stream is valid, we use its velocity prediction directly.

\paratitle{Directional mutual learning.}
We reuse the weights in Eq.~\eqref{eq:fusion_weights} to transfer motion cues between branches during training. For each velocity component, each branch supplies its own prediction as a target for its peer, weighted by its fusion contribution. Let $\bar{k}$ denote the peer branch paired with $k$ and $\operatorname{sg}(\cdot)$ the stop-gradient operator. Over windows in which both streams are valid, the mutual learning objective is
\begin{equation}
\begin{aligned}
\mathcal{L}_{\mathrm{mut}}
&=\mathbb{E}_n\!\left[
\frac{1}{3}\sum_{d=1}^{3}\sum_{k\in\{A,B\}}
\operatorname{sg}(w_{n,d}^{\bar{k}})\,
\ell_{n,d}^{k\leftarrow\bar{k}}
\right],\\
\ell_{n,d}^{k\leftarrow\bar{k}}
&=\left(
\mu_{n,d}^{k}
-\operatorname{sg}(\mu_{n,d}^{\bar{k}})
\right)^2.
\end{aligned}
\label{eq:mutual}
\end{equation}
Applying stop-gradient to both the target and its weight routes each teaching gradient exclusively through the receiving branch, while the reliability heads are trained separately via probabilistic supervision (Sec.~\ref{sec:training}). Because the two weights sum to one, they leave the forward loss value unchanged while governing how the supervisory correction is distributed between branches: when $w_{n,d}^A>w_{n,d}^B$, for example, branch $B$ receives a stronger correction toward $A$ than $A$ receives toward $B$, updating the shared predictor toward the more reliable observation (see the supplement for the explicit gradient derivation).

\begin{table*}[!t]
\centering\small
\caption{\textbf{Dataset overview.} Recording counts and accepted labeled durations (min) are given as training/development-validation. FCU: flight controller; MoCap: motion capture.}
\label{tab:datasets_main1}
\setlength{\tabcolsep}{3pt}
\renewcommand{\arraystretch}{1.05}
\begin{tabular*}{\textwidth}{@{\extracolsep{\fill}}llllrrl@{}}
\toprule
Dataset & Platform & IMU pair & Reference & Recordings & Minutes & Split \\
\midrule
RoNIN~\cite{yan2020ronin} & Human & Phone / Tango & Tango & $65/16$ & $601.27/136.00$ & Recording \\
CoWear~\cite{she2026cowear} & Human & Phone / watch & MoCap glasses & $147/17$ & $189.24/30.06$ & Session \\
GrandTour~\cite{frey2026grandtour} & ANYmal D & ADIS / AlphaSense & DLIO estimate & $33/9$ & $233.14/60.89$ & Location \\
Self-collected & UAV & FCU / Livox & Onboard EKF & $26/7$ & $88.45/20.16$ & Flight group \\
\bottomrule
\end{tabular*}
\smallskip
\begin{minipage}{\textwidth}\footnotesize
CoWear sessions may share participants; its both-valid validation support is $29.23$ min. GrandTour holds out ALB, ARC, and HEAP. Self-collected segments from the same parent flight remain in one partition.
\end{minipage}
\end{table*}

\subsection{Training Objectives and Inference}
\label{sec:training}
Training jointly optimizes single-window accuracy, reliability calibration, and multi-horizon trajectory consistency while separating motion regression from uncertainty fitting via stop-gradient boundaries. Direct velocity supervision $\mathcal{L}_{\mathrm{vel}}=\mathbb{E}_{(n,k)\in\mathcal{V}}[\frac{1}{3}\|\boldsymbol{\mu}_n^k-\mathbf{v}_n^{F,*}\|_2^2]$ anchors both branches to reference motion over valid sensor-window pairs $\mathcal{V}$. Reliability is calibrated via Gaussian negative log-likelihood; up to an additive constant, define
\begin{equation}
\ell(\mathbf{e},\mathbf{u})=\frac{1}{6}\sum_{d=1}^3
\left(\frac{e_d^2}{u_d}+\log u_d\right).
\label{eq:gaussian_objective}
\end{equation}
Over valid pairs $\mathcal{V}$ and windows $\mathcal{U}$ with at least one active sensor, the probabilistic loss is
\begin{equation}
\begin{aligned}
\mathcal{L}_{\mathrm{prob}}={}&\mathbb{E}_{(n,k)\in\mathcal{V}}\,
\ell\!\left(\operatorname{sg}(\boldsymbol{\mu}_n^k)-\mathbf{v}_n^{F,*},
\boldsymbol{\sigma}_n^{k,2}\right)\\
&+\mathbb{E}_{n\in\mathcal{U}}\,
\ell\!\left(\tilde{\mathbf{v}}_n-\mathbf{v}_n^{F,*},\boldsymbol{\sigma}_{f,n}^2\right),
\end{aligned}
\label{eq:probability_loss}
\end{equation}
where $\tilde{\mathbf{v}}_n=\sum_k\mathbf{w}_n^k\odot\operatorname{sg}(\boldsymbol{\mu}_n^k)$. Applying $\operatorname{sg}(\cdot)$ to $\boldsymbol{\mu}_n^k$ isolates reliability calibration from velocity regression: the node-level term fits each branch's marginal variance $\boldsymbol{\sigma}_n^{k,2}$ to its own residual, while the fused term supervises the cross-sensor correlation head $g_\psi$ through $\boldsymbol{\sigma}_{f,n}^2$.

Because window-wise regression can leave low-frequency bias that accumulates into position drift, $\mathcal{L}_{\mathrm{seq}}$ penalizes multi-horizon mean residuals over spans $\mathcal{H}=\{2,5,10\}\,\mathrm{s}$. Since a body-aligned prediction frame $F_n$ rotates continuously during motion, residuals must be expressed in the fixed inertial frame $G$ before temporal averaging. For $M_h$ consecutive windows spanning horizon $h$, let $\bar{\mathbf{e}}_{n,h}^G=M_h^{-1}\sum_{j=0}^{M_h-1}\mathbf{e}_{n+j}^G$, where $\mathbf{e}_{n+j}^G$ denotes the world-frame velocity error of the stop-gradient-weighted fused prediction (rotated into $G$ via $\mathbf{R}_{n+j}^{G\leftarrow F,*}$ when $F_{n+j}\neq G$):
\begin{equation}
\mathcal{L}_{\mathrm{seq}}=\frac{1}{|\mathcal{H}|}\sum_{h\in\mathcal{H}}
\mathbb{E}_{n\in\mathcal{S}_h}\!\left[\frac{1}{3}
\|\bar{\mathbf{e}}_{n,h}^G\|_2^2\right],
\label{eq:temporal_objective}
\end{equation}
where $\mathcal{S}_h$ indexes valid contiguous segments. Combining all objectives yields
\begin{equation}
\mathcal{L}=\mathcal{L}_{\mathrm{vel}}+\mathcal{L}_{\mathrm{prob}}
+\mathcal{L}_{\mathrm{seq}}+\lambda_{\mathrm{mut}}\mathcal{L}_{\mathrm{mut}}.
\label{eq:total_loss}
\end{equation}

To expose the reliability and mutual-learning modules to asymmetric observation quality during training, random single-sensor perturbations are applied to input windows, with $\mathcal{L}_{\mathrm{mut}}$ activated after a supervised warm-up phase once individual velocity and uncertainty estimates stabilize. At inference, only the shared predictor and fusion module are deployed: active branch predictions are combined via Eq.~\eqref{eq:fused_prediction} (or evaluated independently when a single IMU is available), rotated into $G$ using estimated attitudes when required, and integrated at step $\delta t$ to recover trajectories.

%% file: sec/4_Experiments.tex
\section{Experiments}
\label{sec:experiments}

We evaluate \baby across four benchmarks spanning wearable, legged, and aerial platforms to assess both multi-sensor odometry accuracy and standalone single-branch inference. After comparing overall trajectory performance and individual branch predictions against single-IMU baselines (Sec.~\ref{sec:exp_overall}), we dissect the contributions of dual-domain spectral encoding, directional mutual learning, and correlation-aware fusion through controlled ablations (Sec.~\ref{sec:exp_ablation}), followed by analyses of hyperparameter sensitivity, transient sensor corruption, and abrupt stream interruptions (Sec.~\ref{sec:exp_analysis}).

\subsection{Experimental Setup}
\label{sec:exp_setup}
%
\input{tables/overall_multimetric}

\paratitle{Datasets.}
Table~\ref{tab:datasets_main1} summarizes four benchmarks covering wearable, legged, and aerial motion: RoNIN, CoWear, GrandTour, and our self-collected UAV recordings. Each uses a fixed training/development-validation split; reference quality screening, coordinate conventions, and complete partition protocols are detailed in the supplement.

\paratitle{Implementation details.}
Models are trained separately on each dataset using $1$-s, six-axis windows sampled at $200$\,Hz in SI units (retaining gravity without extra filtering) to predict mean velocity over the final $0.2$\,s. Our default training protocol uses AdamW (peak learning rate $6\times 10^{-4}$, weight decay $10^{-4}$) for $33{,}070$ updates, with each batch containing four consecutive $64$-window sequences and a cosine learning-rate schedule after $1\%$ warm-up. Mutual learning starts after $3{,}307$ supervised updates with $\lambda_{\mathrm{mut}}=0.5$ in Eq.~\eqref{eq:total_loss}. We apply model EMA ($0.995$) across random seeds $42$, $3407$, and $2026$, selecting checkpoints by recording-mean ATE (breaking ties with velocity RMSE) while keeping step-zero and warm-up checkpoints eligible. Dataset-selected configurations, baseline objectives, complete multi-seed statistics, and hardware settings are detailed in the supplement.

\paratitle{Baselines.}
We compare against RoNIN~\cite{yan2020ronin}, TLIO-net~\cite{liu2020tlio}, EqNIO-net~\cite{jayanth2025eqnio}, MobileNet~\cite{zeinali2024imunet}, MobileNetV2~\cite{sandler2018mobilenetv2}, TartanIMU~\cite{zhao2025tartanimu}, AirIO-net~\cite{qiu2025airio}, and IMUNet~\cite{zeinali2024imunet}. Single-sensor A/B predictors are trained and selected independently under their registered objectives using a shared IMU-derived integration attitude. Ours (branch) evaluates standalone single-branch predictions (for inputs A and B) with cross-IMU fusion bypassed, whereas Ours (full) reports the fused dual-IMU output. Baseline adaptations and two-predictor fusion controls are detailed in the supplement.

\paratitle{Evaluation metrics.}
We report ATE and $10$-s translational RTE ($\mathrm{RTE}_{10\mathrm{s}}$), averaged equally across recordings. RoNIN uses horizontal XY coordinate RMSE following its official protocol, whereas the remaining datasets use 3D Euclidean RMSE. All methods share identical reference trajectories, segment boundaries, and initial anchors. RoNIN and CoWear are evaluated on the fixed intersection of both valid streams, while GrandTour and Self-collected are evaluated over full accepted timelines.

\subsection{Performance Comparison}
\label{sec:exp_overall}

\begin{figure*}[!t]
\centering
\includegraphics[width=\textwidth]{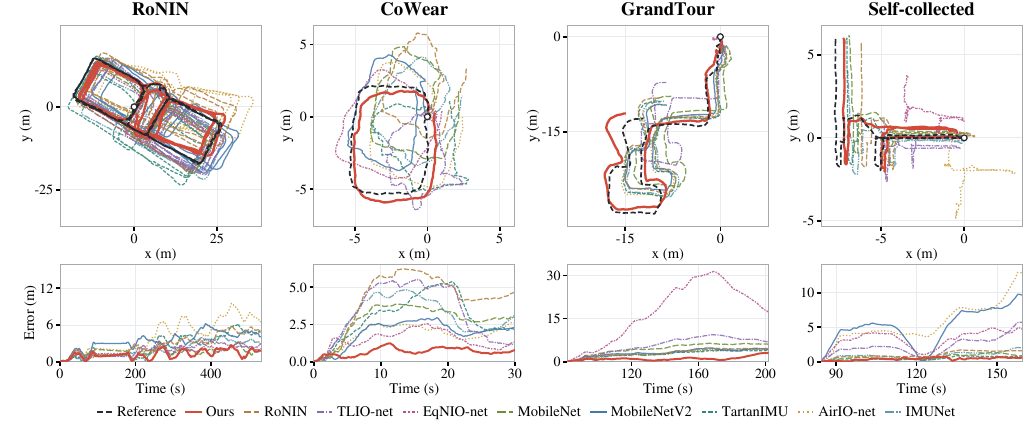}
\caption{\textbf{Trajectory geometry and accumulated drift.} Estimated trajectories (top, using equal XY axis scales and shared starting anchors) and position errors (bottom) across validation recordings from RoNIN, CoWear, GrandTour, and Self-collected.}
\label{fig:qualitative_trajectories}
\end{figure*}

Table~\ref{tab:exp_overall} compares \baby against single-IMU baselines across all four benchmarks. Overall, the fused dual-IMU model (\texttt{Ours (full)}) achieves the lowest ATE and $\mathrm{RTE}_{10\mathrm{s}}$ on every dataset, reaching $3.198$\,m ATE on RoNIN, $3.068$\,m on CoWear, $7.538$\,m on GrandTour, and $2.822$\,m on Self-collected---reducing ATE by $30.5\%$, $17.1\%$, $16.4\%$, and $12.4\%$ relative to the strongest single-input baseline on each benchmark, respectively.

\input{tables/representation_readout_controls}

Notably, standalone single-branch inference (\texttt{Ours (branch)}, $0.797$\,M active parameters) already surpasses every single-IMU baseline across all four datasets even when cross-IMU fusion is bypassed at test time: on RoNIN, branch A achieves $4.215$\,m ATE ($8.4\%$ lower than the best baseline \texttt{AirIO-net-A} at $4.602$\,m), while the stronger standalone branch reduces ATE by $2.2\%$ on CoWear, $1.5\%$ on GrandTour, and $5.9\%$ on Self-collected relative to the top baseline on each dataset. Enabling correlation-aware fusion (\texttt{Ours (full)}) yields a further $6.8\%$--$24.1\%$ ATE reduction over standalone branches by dynamically weighting complementary streams and maintaining continuous velocity estimates across asynchronous dropouts on GrandTour and Self-collected. Qualitatively, \baby preserves tighter loop closure and suppresses long-horizon drift across extended human and robotic trajectories (Fig.~\ref{fig:qualitative_trajectories}).

\subsection{Ablation Study}
\label{sec:exp_ablation}
We evaluate the core architectural and training components of \baby on RoNIN under the unified protocol (Tables~\ref{tab:exp_spectral}--\ref{tab:exp_mutual}).

\input{tables/teaching_controls_panel}

\paratitle{Impact of dual-domain spectral experts.}
Table~\ref{tab:exp_spectral}(a) compares time-domain features, individual wavelet (DWT) and Fourier (FFT) experts, and their combination. Single-domain encoders exhibit clear limitations: \texttt{DWT only} ($4.099\pm 0.208$\,m) shows higher cross-seed variance without global spectral context, whereas \texttt{FFT only} ($3.884\pm 0.116$\,m) lacks localized transient resolution. Combining both domains with equal weights lowers ATE to $3.664\pm 0.119$\,m, outperforming the parameter-matched time-domain baseline ($3.837\pm 0.079$\,m). Replacing equal averaging with our learned Spectral Fusion Gate (\texttt{Adaptive}) further reduces ATE by $12.7\%$ to $3.198\pm 0.165$\,m, confirming the benefit of input-dependent spectral allocation.

\paratitle{Impact of mutual learning.}
\label{sec:exp_mutual}
Table~\ref{tab:exp_mutual} compares no teaching, uniform teaching, and correlation-guided teaching under both fixed-mean and correlation-aware readouts. Even when holding the readout fixed to an unweighted mean, correlation-guided teaching improves predictor accuracy over no teaching ($4.518$\,m vs.\ $4.578$\,m). When paired with correlation-aware fusion (\texttt{Aware}), correlation-guided teaching reaches $3.198\pm 0.165$\,m ATE---a $16.1\%$ reduction over no teaching ($3.810$\,m) and a $14.5\%$ reduction over uniform teaching ($3.741$\,m)---demonstrating strong synergy between reliability-weighted supervision and fusion.

\paratitle{Impact of correlation-aware fusion.}
Table~\ref{tab:exp_fusion}(b) isolates inference-time readout rules on fixed branch predictions. Simple unweighted averaging (\texttt{Mean}, $4.518\pm 0.150$\,m) provides limited gain over the stronger standalone branch (\texttt{IMU-A}, $4.657\pm 0.500$\,m) because it is degraded by the noisier wrist stream (\texttt{IMU-B}, $6.695\pm 0.006$\,m). Modeling marginal variances via inverse-variance weighting (\texttt{Inv.\ var.}) lowers ATE to $3.918\pm 0.048$\,m, while explicitly capturing cross-sensor error correlation (\texttt{Corr.-aware}) reduces ATE by an additional $18.4\%$ to $3.198\pm 0.165$\,m.

\subsection{Further Analysis}
\label{sec:exp_analysis}
\input{sec/sensitivity_analysis}

\paratitle{Robustness to sensor corruption.}
\label{sec:exp_reliability}
To verify that our reliability model identifies transient sensor degradation without online updates, we inject white noise or constant bias at severities $s\in\{0,0.5,1,2,4\}$ for $3$\,s into one preprocessed IMU stream while holding its peer and reference motion fixed. Each of the $16$ RoNIN recordings contributes up to three nonoverlapping $64$-window episodes ($48$ episodes in total), while all recordings remain in global trajectory evaluation without sequence filtering.

\begin{figure}[!t]
\centering
\includegraphics[width=\linewidth]{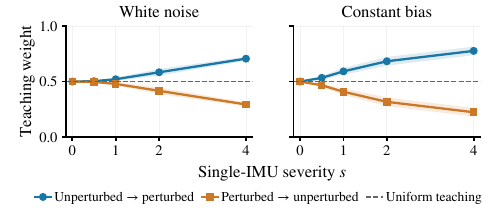}
\caption{\textbf{Reliability under single-stream corruption.} Axis-averaged outgoing teaching weights across $48$ perturbed RoNIN episodes. Shading denotes three-seed sample standard deviation, and dashed lines indicate uniform teaching.}
\label{fig:window_events}
\end{figure}

\begin{figure}[!t]
\centering
\includegraphics[width=\linewidth]{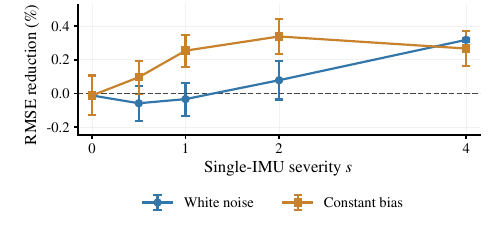}
\caption{\textbf{Prediction gains from directed teaching.} Paired relative velocity RMSE reduction over uniform teaching under fixed-mean readout across $48$ RoNIN corruption episodes. Markers and error bars denote three-seed means and sample standard deviations.}
\label{fig:event_dose}
\end{figure}

As disturbance severity increases, outgoing teaching weights shift monotonically toward the unperturbed peer (Fig.~\ref{fig:window_events}), reaching corrupted/peer ratios of $0.294/0.706$ under white noise and $0.223/0.777$ under constant bias at $s=4$. Figure~\ref{fig:event_dose} complements this allocation response with paired prediction errors under fixed-mean readout: correlation-guided teaching yields consistent positive velocity RMSE reductions across all non-zero severities under constant bias (peaking at $+0.34\%$ at $s=2$ and $+0.27\%$ at $s=4$) and outperforms uniform teaching under strong white noise ($+0.32\%$ at $s=4$). Absolute errors and additional event studies are provided in the supplement.

\paratitle{Robustness to missing streams.}
\label{sec:exp_robustness}
We further evaluate continuity under abrupt sensor dropouts by injecting $1$-s interruptions into one stream while keeping its peer valid, preserving trajectory support and segment anchors. Because both branches estimate a common reference velocity, \baby transitions seamlessly to the surviving branch during the outage and resumes two-stream fusion as soon as complete $1$-s windows reaccumulate, maintaining uninterrupted trajectory integration across failure and recovery without external resets. Extended stress tests covering longer interruptions as well as bias, lag, and stuck-sensor faults are reported in the supplement.

%% file: tables/overall_multimetric.tex
\begin{table*}[t]
\centering
\caption{\textbf{Odometry across human and robotic platforms.} ATE/RTE$_{10\mathrm{s}}$ (m, $\downarrow$) on the four main datasets, using XY coordinate RMSE on RoNIN and 3D Euclidean errors on CoWear, GrandTour, and Self-collected.}
\label{tab:exp_overall}
\begingroup
\fontsize{8.7}{10.4}\selectfont
\setlength{\tabcolsep}{0pt}
\renewcommand{\arraystretch}{0.900}
\definecolor{IOGain}{RGB}{13,66,134}
\definecolor{IOHeatLow}{RGB}{55,126,184}
\definecolor{IOHeatHigh}{RGB}{8,48,107}
\definecolor{IORule}{RGB}{202,207,215}
\renewcommand{\IOgroupRule}{\arrayrulecolor{IORule}\midrule[.22pt]\arrayrulecolor{black}}
\renewcommand{\IOvalue}[1]{\makebox[27pt][r]{#1}}
\renewcommand{\IOdeltaBlank}{\hspace{2pt}\makebox[24pt]{}}
\makeatletter
\newcommand{\IOgainColor}[2]{%
  \dimen0=#2pt\relax
  \def\IOheatBase{IOHeatHigh}\def\IOheatFloor{IOHeatLow}%
  \ifdim\dimen0<0pt
    \dimen0=-\dimen0 \def\IOheatBase{IOLoss}\def\IOheatFloor{IOLoss!75!white}%
  \fi
  \ifdim\dimen0>65pt \dimen0=65pt\fi
  \multiply\dimen0 by100 \divide\dimen0 by65
  \edef\IOheatMix{\strip@pt\dimen0}%
  \color{\IOheatBase!\IOheatMix!\IOheatFloor}%
  \def\IOcurrent{#1}\def\IOstrong{IOFocus}%
  \ifx\IOcurrent\IOstrong\bfseries\fi
}
\makeatother
\renewcommand{\IOdelta}[2]{\hspace{2pt}\raisebox{.2pt}[0pt][0pt]{\makebox[24pt][r]{\fontsize{7}{7.7}\selectfont\IOgainColor{#1}{#2}#2\%}}}
\makeatletter
\newcommand{\IOendRowShade}{\gdef\CT@cell@color{\CT@color{IOPrimary}\@tempdimc=0pt \global\let\CT@cell@color\relax}}
\makeatother
\begin{tabular}{@{}p{.126\linewidth}>{\centering\arraybackslash}p{.055\linewidth}>{\centering\arraybackslash}p{.043\linewidth}
>{\raggedleft\arraybackslash}p{54pt}@{\hspace{3pt}}>{\raggedleft\arraybackslash}p{\dimexpr.194\linewidth-65.25pt\relax}@{\hspace{11pt}}
>{\raggedleft\arraybackslash}p{54pt}@{\hspace{3pt}}>{\raggedleft\arraybackslash}p{\dimexpr.194\linewidth-65.25pt\relax}@{\hspace{11pt}}
>{\raggedleft\arraybackslash}p{54pt}@{\hspace{3pt}}>{\raggedleft\arraybackslash}p{\dimexpr.194\linewidth-65.25pt\relax}@{\hspace{11pt}}
>{\raggedleft\arraybackslash}p{54pt}@{\hspace{3pt}}>{\raggedleft\arraybackslash}p{\dimexpr.194\linewidth-65.25pt\relax}@{}}
\toprule
\multirow{2}{*}{Method} & \multirow{2}{*}{\shortstack{Params\\(M)}} & \multirow{2}{*}{Input} & \multicolumn{2}{c}{\textbf{RoNIN}} & \multicolumn{2}{c}{\textbf{CoWear}} & \multicolumn{2}{c}{\textbf{GrandTour}} & \multicolumn{2}{c}{\textbf{Self-collected}} \\
\cmidrule(lr){4-5}\cmidrule(lr){6-7}\cmidrule(lr){8-9}\cmidrule(lr){10-11}
& & & \multicolumn{1}{c}{ATE ($\Delta$)} & RTE$_{10\mathrm{s}}$ & \multicolumn{1}{c}{ATE ($\Delta$)} & RTE$_{10\mathrm{s}}$ & \multicolumn{1}{c}{ATE ($\Delta$)} & RTE$_{10\mathrm{s}}$ & \multicolumn{1}{c}{ATE ($\Delta$)} & RTE$_{10\mathrm{s}}$ \\
\midrule
\multirow{2}{*}{RoNIN} & \multirow{2}{*}{4.635} & A & \IOvalue{$4.782\IOstd{0.371}$}\IOdelta{IOGain}{+33.1} & \IOvalue{$1.195\IOstd{0.007}$} & \IOvalue{$3.831\IOstd{0.264}$}\IOdelta{IOGain}{+19.9} & \IOvalue{$3.612\IOstd{0.028}$} & \IOvalue{$9.929\IOstd{0.062}$}\IOdelta{IOGain}{+24.1} & \IOvalue{$1.991\IOstd{0.011}$} & \IOvalue{$3.447\IOstd{0.089}$}\IOdelta{IOGain}{+18.1} & \IOvalue{$1.908\IOstd{0.029}$} \\
 &  & B & \IOvalue{$7.034\IOstd{0.049}$}\IOdelta{IOGain}{+54.5} & \IOvalue{$1.397\IOstd{0.105}$} & \IOvalue{$3.797\IOstd{0.199}$}\IOdelta{IOGain}{+19.2} & \IOvalue{$3.820\IOstd{0.070}$} & \IOvalue{$9.114\IOstd{0.142}$}\IOdelta{IOGain}{+17.3} & \IOvalue{$\underline{1.446}\IOstd{0.008}$} & \IOvalue{$3.659\IOstd{0.054}$}\IOdelta{IOGain}{+22.9} & \IOvalue{$2.050\IOstd{0.150}$} \\
\IOgroupRule
\multirow{2}{*}{TLIO-net} & \multirow{2}{*}{5.425} & A & \IOvalue{$4.809\IOstd{0.289}$}\IOdelta{IOGain}{+33.5} & \IOvalue{$1.149\IOstd{0.067}$} & \IOvalue{$3.701\IOstd{0.307}$}\IOdelta{IOFocus}{+17.1} & \IOvalue{$3.675\IOstd{0.056}$} & \IOvalue{$10.118\IOstd{0.457}$}\IOdelta{IOGain}{+25.5} & \IOvalue{$1.987\IOstd{0.120}$} & \IOvalue{$3.364\IOstd{0.027}$}\IOdelta{IOGain}{+16.1} & \IOvalue{$1.998\IOstd{0.030}$} \\
 &  & B & \IOvalue{$6.741\IOstd{0.446}$}\IOdelta{IOGain}{+52.6} & \IOvalue{$1.492\IOstd{0.109}$} & \IOvalue{$3.794\IOstd{0.278}$}\IOdelta{IOGain}{+19.1} & \IOvalue{$3.918\IOstd{0.091}$} & \IOvalue{$9.642\IOstd{0.564}$}\IOdelta{IOGain}{+21.8} & \IOvalue{$1.747\IOstd{0.173}$} & \IOvalue{$3.220\IOstd{0.108}$}\IOdelta{IOFocus}{+12.4} & \IOvalue{$1.869\IOstd{0.032}$} \\
\IOgroupRule
\multirow{2}{*}{EqNIO-net} & \multirow{2}{*}{6.020} & A & \multicolumn{1}{c}{--} & \multicolumn{1}{c}{--} & \IOvalue{$4.276\IOstd{0.171}$}\IOdelta{IOGain}{+28.3} & \IOvalue{$4.074\IOstd{0.179}$} & \IOvalue{$19.665\IOstd{0.075}$}\IOdelta{IOGain}{+61.7} & \IOvalue{$4.429\IOstd{0.001}$} & \IOvalue{$3.886\IOstd{0.181}$}\IOdelta{IOGain}{+27.4} & \IOvalue{$2.021\IOstd{0.040}$} \\
 &  & B & \multicolumn{1}{c}{--} & \multicolumn{1}{c}{--} & \IOvalue{$3.769\IOstd{0.041}$}\IOdelta{IOGain}{+18.6} & \IOvalue{$4.197\IOstd{0.132}$} & \IOvalue{$19.658\IOstd{0.112}$}\IOdelta{IOGain}{+61.7} & \IOvalue{$4.449\IOstd{0.010}$} & \IOvalue{$3.526\IOstd{0.172}$}\IOdelta{IOGain}{+20.0} & \IOvalue{$2.112\IOstd{0.068}$} \\
\IOgroupRule
\multirow{2}{*}{MobileNet} & \multirow{2}{*}{3.180} & A & \IOvalue{$5.076\IOstd{0.181}$}\IOdelta{IOGain}{+37.0} & \IOvalue{$1.121\IOstd{0.023}$} & \IOvalue{$4.067\IOstd{0.278}$}\IOdelta{IOGain}{+24.6} & \IOvalue{$3.600\IOstd{0.074}$} & \IOvalue{$10.858\IOstd{0.249}$}\IOdelta{IOGain}{+30.6} & \IOvalue{$2.020\IOstd{0.023}$} & \IOvalue{$3.937\IOstd{0.084}$}\IOdelta{IOGain}{+28.3} & \IOvalue{$2.127\IOstd{0.017}$} \\
 &  & B & \IOvalue{$8.292\IOstd{0.678}$}\IOdelta{IOGain}{+61.4} & \IOvalue{$1.533\IOstd{0.071}$} & \IOvalue{$3.833\IOstd{0.100}$}\IOdelta{IOGain}{+20.0} & \IOvalue{$3.831\IOstd{0.067}$} & \IOvalue{$10.352\IOstd{0.458}$}\IOdelta{IOGain}{+27.2} & \IOvalue{$1.518\IOstd{0.072}$} & \IOvalue{$4.170\IOstd{0.266}$}\IOdelta{IOGain}{+32.3} & \IOvalue{$2.044\IOstd{0.036}$} \\
\IOgroupRule
\multirow{2}{*}{MobileNetV2} & \multirow{2}{*}{2.185} & A & \IOvalue{$4.910\IOstd{0.309}$}\IOdelta{IOGain}{+34.9} & \IOvalue{$1.143\IOstd{0.021}$} & \IOvalue{$3.944\IOstd{0.095}$}\IOdelta{IOGain}{+22.2} & \IOvalue{$3.707\IOstd{0.071}$} & \IOvalue{$10.832\IOstd{0.184}$}\IOdelta{IOGain}{+30.4} & \IOvalue{$2.033\IOstd{0.014}$} & \IOvalue{$3.824\IOstd{0.000}$}\IOdelta{IOGain}{+26.2} & \IOvalue{$2.183\IOstd{0.000}$} \\
 &  & B & \IOvalue{$7.358\IOstd{0.134}$}\IOdelta{IOGain}{+56.5} & \IOvalue{$1.590\IOstd{0.097}$} & \IOvalue{$3.939\IOstd{0.123}$}\IOdelta{IOGain}{+22.1} & \IOvalue{$3.920\IOstd{0.105}$} & \IOvalue{$9.700\IOstd{0.175}$}\IOdelta{IOGain}{+22.3} & \IOvalue{$1.543\IOstd{0.092}$} & \IOvalue{$3.573\IOstd{0.292}$}\IOdelta{IOGain}{+21.0} & \IOvalue{$2.031\IOstd{0.132}$} \\
\IOgroupRule
\multirow{2}{*}{TartanIMU} & \multirow{2}{*}{5.161} & A & \IOvalue{$4.988\IOstd{0.359}$}\IOdelta{IOGain}{+35.9} & \IOvalue{$1.128\IOstd{0.048}$} & \IOvalue{$3.820\IOstd{0.090}$}\IOdelta{IOGain}{+19.7} & \IOvalue{$3.422\IOstd{0.094}$} & \IOvalue{$10.986\IOstd{0.438}$}\IOdelta{IOGain}{+31.4} & \IOvalue{$2.060\IOstd{0.026}$} & \IOvalue{$3.716\IOstd{0.269}$}\IOdelta{IOGain}{+24.1} & \IOvalue{$1.968\IOstd{0.030}$} \\
 &  & B & \IOvalue{$7.310\IOstd{0.376}$}\IOdelta{IOGain}{+56.3} & \IOvalue{$1.498\IOstd{0.083}$} & \IOvalue{$3.879\IOstd{0.255}$}\IOdelta{IOGain}{+20.9} & \IOvalue{$3.816\IOstd{0.039}$} & \IOvalue{$9.308\IOstd{0.357}$}\IOdelta{IOGain}{+19.0} & \IOvalue{$1.468\IOstd{0.011}$} & \IOvalue{$3.945\IOstd{0.203}$}\IOdelta{IOGain}{+28.5} & \IOvalue{$2.168\IOstd{0.159}$} \\
\IOgroupRule
\multirow{2}{*}{AirIO-net} & \multirow{2}{*}{0.331} & A & \IOvalue{$4.602\IOstd{0.216}$}\IOdelta{IOFocus}{+30.5} & \IOvalue{$\underline{1.049}\IOstd{0.068}$} & \IOvalue{$3.987\IOstd{0.250}$}\IOdelta{IOGain}{+23.0} & \IOvalue{$3.597\IOstd{0.038}$} & \IOvalue{$9.608\IOstd{0.123}$}\IOdelta{IOGain}{+21.5} & \IOvalue{$1.897\IOstd{0.026}$} & \IOvalue{$3.637\IOstd{0.090}$}\IOdelta{IOGain}{+22.4} & \IOvalue{$1.939\IOstd{0.011}$} \\
 &  & B & \IOvalue{$6.910\IOstd{0.048}$}\IOdelta{IOGain}{+53.7} & \IOvalue{$1.341\IOstd{0.028}$} & \IOvalue{$3.943\IOstd{0.079}$}\IOdelta{IOGain}{+22.2} & \IOvalue{$4.065\IOstd{0.064}$} & \IOvalue{$9.016\IOstd{0.113}$}\IOdelta{IOFocus}{+16.4} & \IOvalue{$1.453\IOstd{0.007}$} & \IOvalue{$3.584\IOstd{0.117}$}\IOdelta{IOGain}{+21.3} & \IOvalue{$1.999\IOstd{0.122}$} \\
\IOgroupRule
\multirow{2}{*}{IMUNet} & \multirow{2}{*}{3.663} & A & \IOvalue{$5.108\IOstd{0.214}$}\IOdelta{IOGain}{+37.4} & \IOvalue{$1.069\IOstd{0.003}$} & \IOvalue{$3.774\IOstd{0.475}$}\IOdelta{IOGain}{+18.7} & \IOvalue{$\underline{3.419}\IOstd{0.058}$} & \IOvalue{$10.479\IOstd{0.196}$}\IOdelta{IOGain}{+28.1} & \IOvalue{$2.011\IOstd{0.016}$} & \IOvalue{$3.909\IOstd{0.096}$}\IOdelta{IOGain}{+27.8} & \IOvalue{$2.111\IOstd{0.038}$} \\
 &  & B & \IOvalue{$7.475\IOstd{0.130}$}\IOdelta{IOGain}{+57.2} & \IOvalue{$1.427\IOstd{0.017}$} & \IOvalue{$3.763\IOstd{0.129}$}\IOdelta{IOGain}{+18.5} & \IOvalue{$3.856\IOstd{0.038}$} & \IOvalue{$9.412\IOstd{0.236}$}\IOdelta{IOGain}{+19.9} & \IOvalue{$1.451\IOstd{0.012}$} & \IOvalue{$4.223\IOstd{0.105}$}\IOdelta{IOGain}{+33.2} & \IOvalue{$2.231\IOstd{0.093}$} \\
\midrule[.7pt]
\multirow{2}{*}{Ours (branch)} & \multirow{2}{*}{0.797} & A & \IOBindResult{ronin}{A}{ATE}{\IOvalue{$\underline{4.215}\IOstd{0.122}$}\IOdelta{IOGain}{+24.1}} & \IOBindResult{ronin}{A}{RTE}{\IOvalue{$1.184\IOstd{0.025}$}} & \IOBindResult{cowear}{A}{ATE}{\IOvalue{$\underline{3.618}\IOstd{0.116}$}\IOdelta{IOGain}{+15.2}} & \IOBindResult{cowear}{A}{RTE}{\IOvalue{$3.577\IOstd{0.049}$}} & \IOBindResult{grandtour}{A}{ATE}{\IOvalue{$9.515\IOstd{0.604}$}\IOdelta{IOGain}{+20.8}} & \IOBindResult{grandtour}{A}{RTE}{\IOvalue{$2.034\IOstd{0.028}$}} & \IOBindResult{self}{A}{ATE}{\IOvalue{$3.410\IOstd{0.273}$}\IOdelta{IOGain}{+17.2}} & \IOBindResult{self}{A}{RTE}{\IOvalue{$1.884\IOstd{0.032}$}} \\
 &  & B & \IOBindResult{ronin}{B}{ATE}{\IOvalue{$6.494\IOstd{0.243}$}\IOdelta{IOGain}{+50.8}} & \IOBindResult{ronin}{B}{RTE}{\IOvalue{$1.493\IOstd{0.042}$}} & \IOBindResult{cowear}{B}{ATE}{\IOvalue{$3.625\IOstd{0.168}$}\IOdelta{IOGain}{+15.4}} & \IOBindResult{cowear}{B}{RTE}{\IOvalue{$3.631\IOstd{0.081}$}} & \IOBindResult{grandtour}{B}{ATE}{\IOvalue{$\underline{8.879}\IOstd{0.328}$}\IOdelta{IOGain}{+15.1}} & \IOBindResult{grandtour}{B}{RTE}{\IOvalue{$1.552\IOstd{0.008}$}} & \IOBindResult{self}{B}{ATE}{\IOvalue{$\underline{3.029}\IOstd{0.332}$}\IOdelta{IOGain}{+6.8}} & \IOBindResult{self}{B}{RTE}{\IOvalue{$\mathbf{1.629}\IOstd{0.068}$}} \\
\arrayrulecolor{IOFocus}\midrule[.45pt]\arrayrulecolor{black}
\rowcolor{IOPrimary}[0pt][11pt]
\textcolor{IOFocus}{\textbf{Ours (full)}} & 1.063 & \textcolor{IOFocus}{\textbf{Dual}} & \IOBindResult{ronin}{Dual}{ATE}{\textcolor{IOFocus}{\IOvalue{$\mathbf{3.198}\IOstd{0.165}$}}\IOdeltaBlank} & \IOBindResult{ronin}{Dual}{RTE}{\textcolor{IOFocus}{\IOvalue{$\mathbf{0.914}\IOstd{0.015}$}}} & \IOBindResult{cowear}{Dual}{ATE}{\textcolor{IOFocus}{\IOvalue{$\mathbf{3.068}\IOstd{0.050}$}}\IOdeltaBlank} & \IOBindResult{cowear}{Dual}{RTE}{\textcolor{IOFocus}{\IOvalue{$\mathbf{3.112}\IOstd{0.051}$}}} & \IOBindResult{grandtour}{Dual}{ATE}{\textcolor{IOFocus}{\IOvalue{$\mathbf{7.538}\IOstd{0.368}$}}\IOdeltaBlank} & \IOBindResult{grandtour}{Dual}{RTE}{\textcolor{IOFocus}{\IOvalue{$\mathbf{1.232}\IOstd{0.029}$}}} & \IOBindResult{self}{Dual}{ATE}{\textcolor{IOFocus}{\IOvalue{$\mathbf{2.822}\IOstd{0.062}$}}\IOdeltaBlank} & \IOendRowShade\IOBindResult{self}{Dual}{RTE}{\textcolor{IOFocus}{\IOvalue{$\underline{1.636}\IOstd{0.038}$}}} \\
\bottomrule
\end{tabular}
\endgroup
\smallskip
\begin{minipage}{\linewidth}
\footnotesize\raggedright
Bold/underlined: best/second best. ATE gains: $\Delta=100(1-\mathrm{ATE}_{\mathrm{Ours\,(full)}}/\mathrm{ATE}_{\mathrm{row}})$; blue darkens with gain; bold gains compare with the strongest open-source baseline per dataset.
A/B: input IMU. Branch/full: pre-fusion branch/fused dual-IMU outputs. Params: active network weights (M).
All entries report three-seed means, with sample standard deviations shown where available; complete seed statistics are provided in the supplement. --: not reported.
\end{minipage}
\end{table*}

%% file: tables/representation_readout_controls.tex
\begin{table}[!htbp]
\centering
\caption{\textbf{RoNIN representation and readout controls.} ATE (m), three-seed mean $\pm$ sample standard deviation. Equal/adaptive use DWT + FFT.}
\label{tab:exp_spectral}
\label{tab:exp_fusion}
\begin{minipage}[t]{.54\linewidth}
\centering\small
\setlength{\tabcolsep}{1.5pt}
{\bfseries (a) Representation\par}\smallskip
\begin{tabular*}{\linewidth}{@{\extracolsep{\fill}}lcc@{}}
\toprule
Encoder & ATE $\downarrow$ & Params \\
\midrule
Time-dom.
& $3.837\pm0.079$ & 1.063M \\
DWT only
& $4.099\pm0.208$ & 0.940M \\
FFT only
& $3.884\pm0.116$ & 0.720M \\
Equal
& $3.664\pm0.119$ & 1.046M \\
Adaptive
& \IOResult{ronin}{Dual}{ATE} & 1.063M \\
\bottomrule
\end{tabular*}
\end{minipage}\hfill
\begin{minipage}[t]{.43\linewidth}
\centering\small
\setlength{\tabcolsep}{1.5pt}
{\bfseries (b) Readout\par}\smallskip
\begin{tabular*}{\linewidth}{@{\extracolsep{\fill}}lc@{}}
\toprule
Readout & RoNIN \\
\midrule
IMU-A
& $4.657\pm0.500$ \\
IMU-B
& $6.695\pm0.006$ \\
Mean
& $4.518\pm0.150$ \\
Inv. var.
& $3.918\pm0.048$ \\
Corr.-aware
& \IOResult{ronin}{Dual}{ATE} \\
\bottomrule
\end{tabular*}
\end{minipage}
\end{table}

%% file: tables/teaching_controls_panel.tex
\begin{table}[!htbp]
\centering
\caption{\textbf{RoNIN teaching controls on clean validation.} ATE (m).}
\label{tab:exp_mutual}
\small
\setlength{\tabcolsep}{4pt}
\begin{tabular*}{\linewidth}{@{\extracolsep{\fill}}lcc@{}}
\toprule
Teaching & Mean & Aware \\
\midrule
None
& $4.578\pm0.110$ & $3.810\pm0.116$ \\
Uniform
& $4.524\pm0.127$ & $3.741\pm0.101$ \\
Correlation-guided
& $4.518\pm0.150$ & \IOResult{ronin}{Dual}{ATE} \\
\bottomrule
\end{tabular*}
\end{table}

%% file: sec/sensitivity_analysis.tex
\IfFileExists{figures/mutual_strength.pdf}{%
\paragraph{Hyperparameter sensitivity.}
Figure~\ref{fig:mutual_sensitivity} varies $\lambda_{\mathrm{mut}}$ from $0$ to $1$ on RoNIN with all other settings fixed. Fused ATE falls from $3.342$ to $3.198$\,m at $0.5$, then increases at larger weights.
\begin{figure}[!htbp]
\centering
\includegraphics[width=\linewidth]{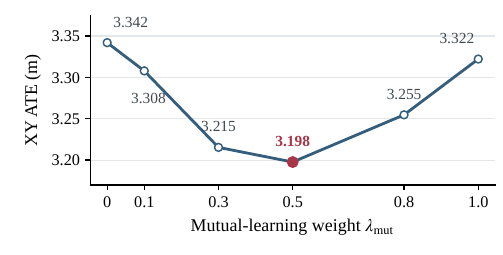}
\caption{\textbf{Effect of mutual-learning strength on RoNIN.} Fused XY ATE ($\downarrow$) at six coefficients with fixed seed $42$. Red marks the lowest observed error.}
\label{fig:mutual_sensitivity}
\end{figure}
}{}

%% file: sec/5_Conclusion.tex
\section{Conclusion}
\label{sec:conclusion}
In this work, we present \baby, a dual-IMU odometry framework that learns from paired inertial observations while allowing either IMU to estimate body velocity independently. Within each branch, wavelet and Fourier experts capture local motion changes and the spectral structure of the full observation window. Our central contribution is to use each branch's role in velocity fusion to determine its influence as a teacher. By accounting for both predictive uncertainty and cross-sensor error correlation, \baby assigns a fusion weight to each estimate for each velocity component and reuses that weight to guide mutual learning. Experiments spanning wearable, aerial, and quadruped motion show that the fusion weights adapt to disturbances affecting individual sensors. Single-branch evaluations further show that directional teaching can improve independent velocity estimation, with the extent of the gain varying across motion conditions.

%% file: sec/6_llm_usage.tex
\section*{LLM Usage Statement}
OpenAI Codex assisted with repository inspection, experiment design and implementation, literature cross-checking, mathematical consistency checks, and the organization and language revision of the Related Work, Method, and Experiments sections. Numerical results were obtained from recorded experiments and checked by recomputing their aggregate statistics. The authors remain responsible for the submission.